\documentclass[sigconf]{acmart}

\AtBeginDocument{%
  \providecommand\BibTeX{{%
    \normalfont B\kern-0.5em{\scshape i\kern-0.25em b}\kern-0.8em\TeX}}}

\copyrightyear{2023}
\acmYear{2023}
\acmDOI{XXXXXXX.XXXXXXX}

\acmConference[The Second In2Writing Workshop @ CHI '23]{April 23, 2023}{Hamburg, Germany}
\usepackage{todonotes}

\setcopyright{none}
\renewcommand\footnotetextcopyrightpermission[1]{}
\begin{document}

\title{Writing Assistants Should Model Social Factors of Language}


\author{Vivek Kulkarni}
\affiliation{
  \institution{Grammarly}
  \city{San Francisco}
  \state{CA}
  \country{USA}
}
\email{vivek.kulkarni@grammarly.com}

\author{Vipul Raheja}
\affiliation{
  \institution{Grammarly}
  \city{San Francisco}
  \state{CA}
  \country{USA}
}
\email{vipul.raheja@grammarly.com}

\renewcommand{\shortauthors}{Kulkarni \& Raheja}

\begin{abstract}
Intelligent writing assistants powered by large language models (LLMs) are more popular today than ever before, but their further
widespread adoption is precluded by sub-optimal performance. In this position paper, we argue that a major reason for this sub-optimal
performance and adoption is a singular focus on the information content of language while ignoring its social aspects. We analyze the
different dimensions of these social factors in the context of writing assistants and propose their incorporation into building smarter,
more effective, and truly personalized writing assistants that would enrich the user experience and contribute to increased user adoption.
\end{abstract}




\maketitle

\section{Introduction}
Advancements in large language models (LLMs) have accelerated their use in many writing assistants \cite{10.1145/3491102.3502030, schick2023peer, du-etal-2022-read, in2writing-2022-intelligent, kim-etal-2022-improving, 10.1145/3490099.3511105}. Millions of people now use AI-driven writing assistants to correct grammar, seek recommendations on word choice, set the right tone, and improve the conciseness and clarity of written content.
However, despite such popularity, it is evident that their sub-optimal performance inhibits further widespread adoption. We argue that an important reason for this sub-optimal performance is due to a limiting modeling assumption -- namely, viewing language as a sequence of tokens with information content. However, as noted in socio-linguistics research and reinforced recently by \citet{hovy-yang-2021-importance}, language is also a social construct used to achieve communicative goals, is grounded in the real world, and is influenced by social aspects. Because underlying social factors heavily influence our understanding of language, they argue that NLP applications should account for social aspects of language to unlock their full potential. They propose a taxonomy of social factors to help researchers reason about these aspects in specific applications. Here, we leverage their proposed taxonomy to comprehensively reason about the various social factors that would specifically benefit intelligent writing assistants, which we outline next.

\section{Applicability of Social Factors of Language in Writing Assistants}
\begin{enumerate}
    \item \textsc{\textbf{Demographics}} \textbf{(Speaker and Receiver Context)}: 
    Prior research has noted age \cite{eckert2017age, tagliamonte2011variationist,barbieri2008patterns, johannsen2015cross}, gender \cite{Holmes1997WomenLA, rickford2013girlz}, and race \cite{blodgett-etal-2016-demographic, blodgett-etal-2018-twitter} influence language use. 
    For example, \citet{tagliamonte2011variationist} observes that word choice is influenced by the age of interlocutors and noted that older people prefer to use ``ha-ha'' over ``lol'' on an instant messenger platform. 
    Further, \citet{van2021sentence} show that there is a gradual deterioration of the ability to interpret long and complex sentences as people age. 
    Similarly, \citet{Johannsen2015CrosslingualSV} report several age and gender-specific variations in word choice and syntactic dependency structures. 
    Finally, \citet{green_2002, jones2015toward} note that African American Vernacular English (AAVE) is a socio-linguistic variety of Standard American English, with distinct syntactic, semantic, and lexical patterns. 
    By modeling these demographic factors, writing assistants can thus improve their recommendations on word choice and sentence phrasing, while seeking to ensure that no biases or stereotypes are perpetuated. \\
    \item \textsc{\textbf{Personality}} \textbf{(Speaker and Receiver Context)}: Personality traits are yet another socio-linguistic variable that significantly influences language use. \citet{schwartz2013personality} reveals significant variation in word use based on latent personality factors. 
    For example, extroverts were more likely to mention social words such as ‘party’, etc. 
    Capturing linguistic variation due to personality factors can make writing assistants truly personalized and account for individual preferences while providing word-choice and sentence phrasing recommendations. \\
    
    \item \textsc{\textbf{Social Relations and Norms}} \textbf{(Social Relation)}: The social relationship between 
    interlocutors is a very important factor that influences language use. Word-choice, tone, sentence structure of communication between two close friends differs significantly from those between colleagues or acquaintances. Many socio-linguistic phenomena might thus manifest based on the social relation. Examples of such socio-linguistic phenomena include the usage of honorifics, slang, code-switching, code-mixing, and avoidance speech. As a use-case, email communications with a close friend might skip all greetings and use slang, while on the other hand, emails to an executive would typically have greetings, appropriate honorifics, and avoid slang. Thus, writing assistants must account for social relations and societal norms. \\
    \item \textsc{\textbf{Time, Geography, and Domain}} 
    Language also demonstrates variation (both syntactic and semantic) across time, geography, domains, and the broader situational context. \cite{Kulkarni2014StatisticallySD, Kulkarni2016FreshmanOF}  Meanings of words can change across all of these dimensions. For example, the word \emph{awesome} had a negative sentiment (inspiring fear) in the 16th century but has taken on its positive sense over time. Similarly, different tokens may be used to refer to the same real-world concept (\textit{zucchini} in the US vs \textit{courgette} in the UK) \cite{Kulkarni2016FreshmanOF}. Writing assistants not accounting for such linguistic variation may lead to poor user experience (e.g., incorrect sentiment or tone detection, or word recommendations). \\
    \item \textsc{\textbf{Intent}} \textbf{(Communicative Goal)}: Writing assistants need to have an intimate knowledge of the communicative intent of the user to be effective. Recommendations on word choice, sentence and paragraph restructuring, and feedback on sentiment and tone depend on the user's specific communicative goal (which might be to inform, entertain, persuade, or narrate) and targeted setting (academic, creative writing,  or conversational). For example, in content targeted for an academic publication, writing assistants might assist users by recommending templates and phrases that seek to achieve specific communicative goals like (a) introducing standard views, quotations, and an ongoing debate, (b) contrasting with prior work, and (c) motivating claims. 
\end{enumerate}

\section{Closing Remarks}
In this paper,  we discuss clear use cases of intelligent writing assistants that would benefit by adopting a richer view of language, which accounts for its social aspects. Building writing assistants that adopt this richer view of language opens up exciting research directions. First, a majority of the current evaluation benchmarks used for evaluating writing assistants today ignore these social factors. Therefore, there is a critical need to construct comprehensive evaluation benchmarks grounded in social factors. Second, note that many of these social factors are extra-linguistic and may involve modeling multiple modalities. Research needs to be undertaken around exploring approaches to modeling these social factors in a manner that is best suited toward their incorporation in writing assistants. Finally,  one needs to work within appropriate considerations around data/user privacy and ethics to ensure models benefit end users and not perpetuate negative biases. We thus conclude by urging the community to advance further research on the social aspects of language and how these aspects can relate to building smarter, more effective, highly personalized, and inclusive writing assistants. 

\bibliographystyle{ACM-Reference-Format}
\bibliography{sample-base}


\begin{thebibliography}{22}


\ifx \showCODEN    \undefined \def \showCODEN     #1{\unskip}     \fi
\ifx \showDOI      \undefined \def \showDOI       #1{#1}\fi
\ifx \showISBNx    \undefined \def \showISBNx     #1{\unskip}     \fi
\ifx \showISBNxiii \undefined \def \showISBNxiii  #1{\unskip}     \fi
\ifx \showISSN     \undefined \def \showISSN      #1{\unskip}     \fi
\ifx \showLCCN     \undefined \def \showLCCN      #1{\unskip}     \fi
\ifx \shownote     \undefined \def \shownote      #1{#1}          \fi
\ifx \showarticletitle \undefined \def \showarticletitle #1{#1}   \fi
\ifx \showURL      \undefined \def \showURL       {\relax}        \fi
\providecommand\bibfield[2]{#2}
\providecommand\bibinfo[2]{#2}
\providecommand\natexlab[1]{#1}
\providecommand\showeprint[2][]{arXiv:#2}

\bibitem[Barbieri(2008)]%
        {barbieri2008patterns}
\bibfield{author}{\bibinfo{person}{Federica Barbieri}.}
  \bibinfo{year}{2008}\natexlab{}.
\newblock \showarticletitle{Patterns of age-based linguistic variation in
  American English 1}.
\newblock \bibinfo{journal}{\emph{Journal of sociolinguistics}}
  \bibinfo{volume}{12}, \bibinfo{number}{1} (\bibinfo{year}{2008}),
  \bibinfo{pages}{58--88}.
\newblock


\bibitem[Blodgett et~al\mbox{.}(2016)]%
        {blodgett-etal-2016-demographic}
\bibfield{author}{\bibinfo{person}{Su~Lin Blodgett}, \bibinfo{person}{Lisa
  Green}, {and} \bibinfo{person}{Brendan O{'}Connor}.}
  \bibinfo{year}{2016}\natexlab{}.
\newblock \showarticletitle{Demographic Dialectal Variation in Social Media: A
  Case Study of {A}frican-{A}merican {E}nglish}. In
  \bibinfo{booktitle}{\emph{Proceedings of the 2016 Conference on Empirical
  Methods in Natural Language Processing}}. \bibinfo{publisher}{Association for
  Computational Linguistics}, \bibinfo{address}{Austin, Texas},
  \bibinfo{pages}{1119--1130}.
\newblock
\urldef\tempurl%
\url{https://doi.org/10.18653/v1/D16-1120}
\showDOI{\tempurl}


\bibitem[Blodgett et~al\mbox{.}(2018)]%
        {blodgett-etal-2018-twitter}
\bibfield{author}{\bibinfo{person}{Su~Lin Blodgett}, \bibinfo{person}{Johnny
  Wei}, {and} \bibinfo{person}{Brendan O{'}Connor}.}
  \bibinfo{year}{2018}\natexlab{}.
\newblock \showarticletitle{{T}witter {U}niversal {D}ependency Parsing for
  {A}frican-{A}merican and Mainstream {A}merican {E}nglish}. In
  \bibinfo{booktitle}{\emph{Proceedings of the 56th Annual Meeting of the
  Association for Computational Linguistics (Volume 1: Long Papers)}}.
  \bibinfo{publisher}{Association for Computational Linguistics},
  \bibinfo{address}{Melbourne, Australia}, \bibinfo{pages}{1415--1425}.
\newblock
\urldef\tempurl%
\url{https://doi.org/10.18653/v1/P18-1131}
\showDOI{\tempurl}


\bibitem[Du et~al\mbox{.}(2022)]%
        {du-etal-2022-read}
\bibfield{author}{\bibinfo{person}{Wanyu Du}, \bibinfo{person}{Zae~Myung Kim},
  \bibinfo{person}{Vipul Raheja}, \bibinfo{person}{Dhruv Kumar}, {and}
  \bibinfo{person}{Dongyeop Kang}.} \bibinfo{year}{2022}\natexlab{}.
\newblock \showarticletitle{Read, Revise, Repeat: A System Demonstration for
  Human-in-the-loop Iterative Text Revision}. In
  \bibinfo{booktitle}{\emph{Proceedings of the First Workshop on Intelligent
  and Interactive Writing Assistants (In2Writing 2022)}}.
  \bibinfo{publisher}{Association for Computational Linguistics},
  \bibinfo{address}{Dublin, Ireland}, \bibinfo{pages}{96--108}.
\newblock
\urldef\tempurl%
\url{https://doi.org/10.18653/v1/2022.in2writing-1.14}
\showDOI{\tempurl}


\bibitem[Eckert(2017)]%
        {eckert2017age}
\bibfield{author}{\bibinfo{person}{Penelope Eckert}.}
  \bibinfo{year}{2017}\natexlab{}.
\newblock \showarticletitle{Age as a sociolinguistic variable}.
\newblock \bibinfo{journal}{\emph{The handbook of sociolinguistics}}
  (\bibinfo{year}{2017}), \bibinfo{pages}{151--167}.
\newblock


\bibitem[Green(2002)]%
        {green_2002}
\bibfield{author}{\bibinfo{person}{Lisa~J. Green}.}
  \bibinfo{year}{2002}\natexlab{}.
\newblock \bibinfo{booktitle}{\emph{African American English: A Linguistic
  Introduction}}.
\newblock \bibinfo{publisher}{Cambridge University Press}.
\newblock
\urldef\tempurl%
\url{https://doi.org/10.1017/CBO9780511800306}
\showDOI{\tempurl}


\bibitem[Holmes(1997)]%
        {Holmes1997WomenLA}
\bibfield{author}{\bibinfo{person}{Janet Holmes}.}
  \bibinfo{year}{1997}\natexlab{}.
\newblock \showarticletitle{Women, Language and Identity}.
\newblock \bibinfo{journal}{\emph{Journal of Sociolinguistics}}
  \bibinfo{volume}{1} (\bibinfo{year}{1997}), \bibinfo{pages}{195--223}.
\newblock


\bibitem[Hovy and Yang(2021)]%
        {hovy-yang-2021-importance}
\bibfield{author}{\bibinfo{person}{Dirk Hovy} {and} \bibinfo{person}{Diyi
  Yang}.} \bibinfo{year}{2021}\natexlab{}.
\newblock \showarticletitle{The Importance of Modeling Social Factors of
  Language: Theory and Practice}. In \bibinfo{booktitle}{\emph{Proceedings of
  the 2021 Conference of the North American Chapter of the Association for
  Computational Linguistics: Human Language Technologies}}.
  \bibinfo{publisher}{Association for Computational Linguistics},
  \bibinfo{address}{Online}, \bibinfo{pages}{588--602}.
\newblock
\urldef\tempurl%
\url{https://doi.org/10.18653/v1/2021.naacl-main.49}
\showDOI{\tempurl}


\bibitem[Huang et~al\mbox{.}(2022)]%
        {in2writing-2022-intelligent}
\bibfield{editor}{\bibinfo{person}{Ting-Hao~'Kenneth' Huang},
  \bibinfo{person}{Vipul Raheja}, \bibinfo{person}{Dongyeop Kang},
  \bibinfo{person}{John Joon~Young Chung}, \bibinfo{person}{Daniel Gissin},
  \bibinfo{person}{Mina Lee}, {and} \bibinfo{person}{Katy~Ilonka Gero}} (Eds.).
  \bibinfo{year}{2022}\natexlab{}.
\newblock \bibinfo{booktitle}{\emph{Proceedings of the First Workshop on
  Intelligent and Interactive Writing Assistants (In2Writing 2022)}}.
  \bibinfo{publisher}{Association for Computational Linguistics},
  \bibinfo{address}{Dublin, Ireland}.
\newblock
\urldef\tempurl%
\url{https://aclanthology.org/2022.in2writing-1.0}
\showURL{%
\tempurl}


\bibitem[Johannsen et~al\mbox{.}(2015a)]%
        {johannsen2015cross}
\bibfield{author}{\bibinfo{person}{Anders Johannsen}, \bibinfo{person}{Dirk
  Hovy}, {and} \bibinfo{person}{Anders S{\o}gaard}.}
  \bibinfo{year}{2015}\natexlab{a}.
\newblock \showarticletitle{Cross-lingual syntactic variation over age and
  gender}. In \bibinfo{booktitle}{\emph{Proceedings of the nineteenth
  conference on computational natural language learning}}.
  \bibinfo{pages}{103--112}.
\newblock


\bibitem[Johannsen et~al\mbox{.}(2015b)]%
        {Johannsen2015CrosslingualSV}
\bibfield{author}{\bibinfo{person}{Anders Johannsen}, \bibinfo{person}{Dirk
  Hovy}, {and} \bibinfo{person}{Anders S{\o}gaard}.}
  \bibinfo{year}{2015}\natexlab{b}.
\newblock \showarticletitle{Cross-lingual syntactic variation over age and
  gender}. In \bibinfo{booktitle}{\emph{Conference on Computational Natural
  Language Learning}}.
\newblock


\bibitem[Jones(2015)]%
        {jones2015toward}
\bibfield{author}{\bibinfo{person}{Taylor Jones}.}
  \bibinfo{year}{2015}\natexlab{}.
\newblock \showarticletitle{Toward a description of african american vernacular
  english dialect regions using “black twitter”}.
\newblock \bibinfo{journal}{\emph{American Speech}} \bibinfo{volume}{90},
  \bibinfo{number}{4} (\bibinfo{year}{2015}), \bibinfo{pages}{403--440}.
\newblock


\bibitem[Kim et~al\mbox{.}(2022)]%
        {kim-etal-2022-improving}
\bibfield{author}{\bibinfo{person}{Zae~Myung Kim}, \bibinfo{person}{Wanyu Du},
  \bibinfo{person}{Vipul Raheja}, \bibinfo{person}{Dhruv Kumar}, {and}
  \bibinfo{person}{Dongyeop Kang}.} \bibinfo{year}{2022}\natexlab{}.
\newblock \showarticletitle{Improving Iterative Text Revision by Learning Where
  to Edit from Other Revision Tasks}. In \bibinfo{booktitle}{\emph{Proceedings
  of the 2022 Conference on Empirical Methods in Natural Language Processing}}.
  \bibinfo{publisher}{Association for Computational Linguistics},
  \bibinfo{address}{Abu Dhabi, United Arab Emirates},
  \bibinfo{pages}{9986--9999}.
\newblock
\urldef\tempurl%
\url{https://aclanthology.org/2022.emnlp-main.678}
\showURL{%
\tempurl}


\bibitem[Kulkarni et~al\mbox{.}(2014)]%
        {Kulkarni2014StatisticallySD}
\bibfield{author}{\bibinfo{person}{Vivek Kulkarni}, \bibinfo{person}{Rami
  Al-Rfou}, \bibinfo{person}{Bryan Perozzi}, {and} \bibinfo{person}{Steven
  Skiena}.} \bibinfo{year}{2014}\natexlab{}.
\newblock \showarticletitle{Statistically Significant Detection of Linguistic
  Change}.
\newblock \bibinfo{journal}{\emph{Proceedings of the 24th International
  Conference on World Wide Web}} (\bibinfo{year}{2014}).
\newblock


\bibitem[Kulkarni et~al\mbox{.}(2016)]%
        {Kulkarni2016FreshmanOF}
\bibfield{author}{\bibinfo{person}{Vivek Kulkarni}, \bibinfo{person}{Bryan
  Perozzi}, {and} \bibinfo{person}{Steven Skiena}.}
  \bibinfo{year}{2016}\natexlab{}.
\newblock \showarticletitle{Freshman or Fresher? Quantifying the Geographic
  Variation of Language in Online Social Media}. In
  \bibinfo{booktitle}{\emph{International Conference on Web and Social Media}}.
\newblock


\bibitem[Lee et~al\mbox{.}(2022)]%
        {10.1145/3491102.3502030}
\bibfield{author}{\bibinfo{person}{Mina Lee}, \bibinfo{person}{Percy Liang},
  {and} \bibinfo{person}{Qian Yang}.} \bibinfo{year}{2022}\natexlab{}.
\newblock \showarticletitle{CoAuthor: Designing a Human-AI Collaborative
  Writing Dataset for Exploring Language Model Capabilities}. In
  \bibinfo{booktitle}{\emph{Proceedings of the 2022 CHI Conference on Human
  Factors in Computing Systems}} (New Orleans, LA, USA)
  \emph{(\bibinfo{series}{CHI '22})}. \bibinfo{publisher}{Association for
  Computing Machinery}, \bibinfo{address}{New York, NY, USA}, Article
  \bibinfo{articleno}{388}, \bibinfo{numpages}{19}~pages.
\newblock
\showISBNx{9781450391573}
\urldef\tempurl%
\url{https://doi.org/10.1145/3491102.3502030}
\showDOI{\tempurl}


\bibitem[Rickford and Price(2013)]%
        {rickford2013girlz}
\bibfield{author}{\bibinfo{person}{John Rickford} {and}
  \bibinfo{person}{Mackenzie Price}.} \bibinfo{year}{2013}\natexlab{}.
\newblock \showarticletitle{Girlz II women: Age-grading, language change and
  stylistic variation}.
\newblock \bibinfo{journal}{\emph{Journal of Sociolinguistics}}
  \bibinfo{volume}{17}, \bibinfo{number}{2} (\bibinfo{year}{2013}),
  \bibinfo{pages}{143--179}.
\newblock


\bibitem[Schick et~al\mbox{.}(2023)]%
        {schick2023peer}
\bibfield{author}{\bibinfo{person}{Timo Schick}, \bibinfo{person}{Jane~A. Yu},
  \bibinfo{person}{Zhengbao Jiang}, \bibinfo{person}{Fabio Petroni},
  \bibinfo{person}{Patrick Lewis}, \bibinfo{person}{Gautier Izacard},
  \bibinfo{person}{Qingfei You}, \bibinfo{person}{Christoforos Nalmpantis},
  \bibinfo{person}{Edouard Grave}, {and} \bibinfo{person}{Sebastian Riedel}.}
  \bibinfo{year}{2023}\natexlab{}.
\newblock \showarticletitle{{PEER}: A Collaborative Language Model}. In
  \bibinfo{booktitle}{\emph{International Conference on Learning
  Representations}}.
\newblock
\urldef\tempurl%
\url{https://openreview.net/forum?id=KbYevcLjnc}
\showURL{%
\tempurl}


\bibitem[Schwartz et~al\mbox{.}(2013)]%
        {schwartz2013personality}
\bibfield{author}{\bibinfo{person}{H~Andrew Schwartz},
  \bibinfo{person}{Johannes~C Eichstaedt}, \bibinfo{person}{Margaret~L Kern},
  \bibinfo{person}{Lukasz Dziurzynski}, \bibinfo{person}{Stephanie~M Ramones},
  \bibinfo{person}{Megha Agrawal}, \bibinfo{person}{Achal Shah},
  \bibinfo{person}{Michal Kosinski}, \bibinfo{person}{David Stillwell},
  \bibinfo{person}{Martin~EP Seligman}, {et~al\mbox{.}}}
  \bibinfo{year}{2013}\natexlab{}.
\newblock \showarticletitle{Personality, gender, and age in the language of
  social media: The open-vocabulary approach}.
\newblock \bibinfo{journal}{\emph{PloS one}} \bibinfo{volume}{8},
  \bibinfo{number}{9} (\bibinfo{year}{2013}), \bibinfo{pages}{e73791}.
\newblock


\bibitem[Tagliamonte(2011)]%
        {tagliamonte2011variationist}
\bibfield{author}{\bibinfo{person}{Sali~A Tagliamonte}.}
  \bibinfo{year}{2011}\natexlab{}.
\newblock \bibinfo{booktitle}{\emph{Variationist sociolinguistics: Change,
  observation, interpretation}}.
\newblock \bibinfo{publisher}{John Wiley \& Sons}.
\newblock


\bibitem[van Boxtel and Lawyer(2021)]%
        {van2021sentence}
\bibfield{author}{\bibinfo{person}{Willem van Boxtel} {and}
  \bibinfo{person}{Laurel Lawyer}.} \bibinfo{year}{2021}\natexlab{}.
\newblock \showarticletitle{Sentence comprehension in ageing and Alzheimer's
  disease}.
\newblock \bibinfo{journal}{\emph{Language and Linguistics Compass}}
  \bibinfo{volume}{15}, \bibinfo{number}{6} (\bibinfo{year}{2021}),
  \bibinfo{pages}{e12430}.
\newblock


\bibitem[Yuan et~al\mbox{.}(2022)]%
        {10.1145/3490099.3511105}
\bibfield{author}{\bibinfo{person}{Ann Yuan}, \bibinfo{person}{Andy Coenen},
  \bibinfo{person}{Emily Reif}, {and} \bibinfo{person}{Daphne Ippolito}.}
  \bibinfo{year}{2022}\natexlab{}.
\newblock \showarticletitle{Wordcraft: Story Writing With Large Language
  Models}. In \bibinfo{booktitle}{\emph{27th International Conference on
  Intelligent User Interfaces}} (Helsinki, Finland) \emph{(\bibinfo{series}{IUI
  '22})}. \bibinfo{publisher}{Association for Computing Machinery},
  \bibinfo{address}{New York, NY, USA}, \bibinfo{pages}{841–852}.
\newblock
\showISBNx{9781450391443}
\urldef\tempurl%
\url{https://doi.org/10.1145/3490099.3511105}
\showDOI{\tempurl}


\end{thebibliography}

\appendix

\end{document}